\documentclass[runningheads]{llncs}

\usepackage[T1]{fontenc}
\usepackage{graphicx}
\usepackage{multirow}
\usepackage{booktabs}
\usepackage{xcolor}
\usepackage{balance}
\usepackage{subcaption}
\usepackage{url}

\definecolor{boxgrey}{HTML}{F3F3F3}
\newcommand{\hlbox}[2]{
  \begin{center}
    \fcolorbox{white}{boxgrey}{
      \parbox{.95\columnwidth}{\noindent \textbf{#1}. \textit{#2}}
    }
  \end{center}
}
\begin{document}

\title{A Structured Unplugged Approach for Foundational AI Literacy in Primary Education}
\titlerunning{Foundational AI Literacy in Primary Education}

\author{Maria Cristina Carrisi\inst{1}\orcidID{0000-0002-2837-3971} \and
Mirko Marras\inst{1}\orcidID{0000-0003-1989-6057} \and
Sara Vergallo\inst{2}\orcidID{0009-0006-2129-5583}}

\authorrunning{M.C. Carrisi et al.}

\institute{
University of Cagliari, Cagliari, Italy\\
\email{\{mariacri.carrisi,mirko.marras\}@unica.it}
\and
University of Macerata, Macerata, Italy\\
\email{s.vergallo@unimc.it}
}

\maketitle            

\begin{abstract}
Younger generations are growing up in a world increasingly shaped by intelligent technologies, making early AI literacy crucial for developing the skills to critically understand and navigate them. However, education in this field often emphasizes tool-based learning, prioritizing usage over understanding the underlying concepts. This lack of knowledge leaves non-experts, especially children, prone to misconceptions, unrealistic expectations, and difficulties in recognizing biases and stereotypes. In this paper, we propose a structured and replicable teaching approach that fosters foundational AI literacy in primary students, by building upon core mathematical elements closely connected to and of interest in primary curricula, 
to strengthen conceptualization, data representation, classification reasoning, and evaluation of AI. To assess the effectiveness of our approach, we conducted an empirical study with thirty-one fifth-grade students across two classes, evaluating their progress through a post-test and a satisfaction survey. Our results indicate improvements in terminology understanding and usage, features description, logical reasoning, and evaluative skills, with students showing a deeper comprehension of decision-making processes and their limitations. Moreover, the approach proved engaging, with students particularly enjoying activities that linked AI concepts to real-world reasoning. \textbf{Materials}: \url{https://github.com/tail-unica/ai-literacy-primary-ed}.
\end{abstract}

\section{Introduction}

\noindent \textbf{Motivation}. Artificial Intelligence (AI) is now embedded in everyday life, shaping how individuals interact with technology and process information. Children, in particular, encounter AI-powered systems daily, when using voice assistants, watching recommended videos, or playing adaptive educational games, as examples. However, without a foundational understanding of AI, they risk passively interacting 
with these technologies, leading to misconceptions, unrealistic expectations, and inability to assess algorithmic outputs \cite{Williams,Yang}. For example, a child might trust a chatbot historical explanation without questioning its accuracy. Early AI literacy becomes thus crucial for enabling young learners to comprehend how such systems function, be aware that they are limited systems, recognize such limitations, and evaluate their implications. This is crucial to raise responsible citizens of the future who use technology with awareness \cite{UNICEF}.

AI literacy extends beyond technology use; it requires and reinforces 
fundamental mathematical concepts. 
AI models, for instance, operate on principles similar to sorting, helping students understand how objects can be grouped based on shared features, like categorizing personal expenses into needs. Likewise, AI systems rely on measures of accuracy to assess their performance, similar to how students would track their progress in physical activities by comparing running times or steps taken each day. AI literacy also strengthens logical progression by encouraging students to follow sequences, recognize patterns, and check their conclusions, as they would do when following a set of instructions for a project. By developing AI literacy, children not only become critical AI users but also build strong problem-solving and analytical skills and enforce math knowledge. 


\noindent \textbf{Prior Works}.
AI has been a subject of study since the mid-20th century, with early discussions emphasizing the need to structure its foundational studies. The pedagogical implications of AI education were considered since early 1970s~\cite{Papert}, showing the importance of integrating AI and CS into education from childhood. However, limited computational resources historically slowed down both AI development and its integration into curricula. In recent years, AI and robotics have renewed interest in early AI education, prompting organizations to establish initiatives, e.g., \texttt{Digital Education Action Plan}~\cite{EUR}, \texttt{Informatics for All}~\cite{Inf}, and \texttt{National AI Initiative} in the US~\cite{US}. Many countries are starting to reason on including CS into school curricula, often including AI topics.

With this growing interest, the need for structured frameworks has become urgent. For instance, the \texttt{AI4K12} initiative \cite{AI4K12} aimed to define "Five Big Ideas" that every K-12 student should understand about AI~\cite{Touretzky1,Touretzky2}. Numerous tools~\cite{Code,google,ML4kids,Gresse} have been developed to facilitate AI learning, allowing students to experiment with underlying concepts. While offering an accessible entry point into AI, this tools often function as "black boxes", limiting students' understanding of the core mechanisms~\cite{Van}. This raises concerns about whether current AI education fosters true comprehension or just reinforces procedural familiarity.

\noindent \textbf{Open Issues}.  
Studies indicate that exposure to AI tools alone does not necessarily improve conceptual understanding. For instance, using AI-based tools and robots was found not to enhance students’ CS competence more than other activities, nor their awareness of AI’s functioning \cite{Capecchi1}. Similarly, a recent survey~\cite{Sanusi} shows a lack of structured learning paths for AI, with the existing ones often focused on tool usage rather than concepts. While unplugged activities have been proposed to address this gap~\cite{Lindner,Ma,Shamir}, they mainly target older students. 

Therefore, there remains a lack of well-structured curricular proposals for primary school students, particularly on foundational AI concepts. A key challenge for non-expert learners in CS education is the underlying mathematical difficulty~\cite{Sulmont}, which becomes pronounced in data-related topics such as AI. Deficiencies in classification (sets) and data representation (trees, tables)~\cite{Grillenberger} hinder students' ability to engage with AI concepts, though these skills are introduced in primary school. However, they are often insufficiently emphasized in early education~\cite{Grillenberger}. To our knowledge, no learning path systematically integrates AI education with mathematical skill development for primary schools. 

\noindent \textbf{Contributions}. 
In this paper, we explore how primary education can effectively integrate foundational AI concepts with 
mathematics to enhance students' conceptual understanding and engagement. Specifically, we investigate how a structured learning path can bridge AI and mathematical reasoning by emphasizing, for instance, classification, set theory, and data representation. To this end, we designed an unplugged, hands-on curriculum that introduces the theoretical and mathematical foundations of AI through interactive and problem-solving activities. Our research focuses on assessing the impact of this approach on students' AI comprehension (\textbf{RQ$_1$}), its role in strengthening mathematical skills (\textbf{RQ$_2$}), and its effectiveness in fostering engagement and interest (\textbf{RQ$_3$}).

To address these research questions, we make three key contributions. First, we present the design of a novel learning path that systematically integrates AI principles with mathematical concepts, ensuring alignment with primary school curricula. Second, we detail the implementation of this learning path, including the selection and preparation of instructional materials, which were carefully curated to support students' cognitive development in both AI and mathematics. Third, we evaluate the effectiveness of the learning path through a study involving two primary school classes, analyzing both quantitative performance and qualitative feedback to assess conceptual gains and engagement.


\section{The Proposed Learning Path} \label{sec:path}
In this section, we present the design and implementation of a structured learning path aimed at introducing primary school students to key AI concepts while integrating foundational mathematical reasoning. Our approach is designed to provide a coherent and progressive experience. The path is structured to reinforce prior knowledge, guide students in identifying system limitations, and introduce new representation models to support cognitive development. We ground in constructivism and constructionism \cite{Piaget,Papert}, and adopt a spiral learning approach \cite{Bruner}, ensuring that concepts are reintroduced at increasing levels of complexity.

To implement this path, we combine original instructional materials with adapted resources from established educational frameworks. The learning modules are structured following learning-by-doing \cite{Dewey} and learning-by-necessity \cite{Sinha} methodologies. These strategies encourage students to actively experiment, refine their knowledge, and engage in iterative problem-solving, with targeted teacher interventions whenever prior approaches proved insufficient. Additionally, semiotic representation in reasoning \cite{Radford}  facilitates students' understanding of classification concepts via multiple representational models, e.g., Euler-Venn diagrams, tabular data structures, and decision trees. 
Topics and themes are chosen according to four out of the "Five Big Ideas" \cite{AI4K12,Touretzky1},
namely perception, representation and reasoning, learning, and societal impact\footnote{As CS is not generally taught in primary schools in \emph{anonymized country} and many others, we excluded natural interaction to prioritize foundational computational thinking over more advanced human-computer interaction concepts.}. 

\subsection{Module 1: Introduction to AI (2 hours)}
The first module establishes foundations of CS and AI, providing students with the knowledge to comprehend how AI uses data. 
An ice breaking questionnaire allows to monitor preconception on CS and AI and evaluate initial classification and argumentative abilities. Then, the session aims to dismantle common misconceptions, emphasizing AI as a human-engineered tool designed to automate data processing, rather than an autonomous entity possessing intelligence.

The lesson begins with an exploration of CS as a discipline, tracing its origins and highlighting its role in developing computational tools for automated information processing. Some historical illustrate how humans have always strove to optimize their work, helping students to see AI as a continuation of past efforts to enhance automation, and sparking discussions on the societal impact of new technologies on the job market. A key distinction was introduced between \textbf{data} and \textbf{information}. Students are introduced to computers as machines that receive, process, store, and output data based on precisely defined human instructions and not on some intrinsic intelligence. Errors in AI systems originate from human design flaws rather than intrinsic machine faults.

To bridge theoretical concepts with real-world applications and to experience different ways of \textbf{computer perception}, students analyze automation systems, including a supermarket checkout system and two automated irrigation systems. The first illustrates how barcode scanners convert product codes into meaningful data, enabling automated price retrieval and cost calculation. The second compares handmade and industrial moisture sensors to demonstrate how different devices influence data quality. The handmade sensor uses alligator clips and aluminum foil and detects only two states (presence or absence of water), while the industrial sensor provides specific humidity values. The third system uses humidity and light sensors to determine when watering is necessary, following a structured set of predefined logical rules. This helps students understand how \textbf{multiple data sources} enhance perception and how such systems rely on \textbf{rule-based decision-making}.
Subsequently, we retain fundamental that students experience a machine learning based tool. We suggest the first 5 steps of the \textit{AI for the Oceans} activity \cite{Code}. It is designed for autonomous use by children and is accompanied by explanatory videos that can be skipped to guide personally students across the activities, to emphasize some steps (\textbf{train} and \textbf{test}), their ordering and meaning, and to introduce specific terminology with contextualization (\textbf{data}, \textbf{labels}, \textbf{model}, \textbf{rule}). This hands-on exercise allows also to observe how incorrect training (e.g., mislabeling an apple as a fish) leads to inaccurate recognition. In alignment with a learning-by-doing approach, students engage in discussions about AI’s limitations, reinforcing the idea that AI does not "think" independently but operates within structured input-output relationships, ultimately functioning as an extension of human decision-making.

\subsection{Module 2: Classification Principles (2 hours)}
The second module introduces students to classification, focusing on how AI categorizes objects based on predefined rules extracted by data-driven processes. The session follows a spiral learning approach, reinforcing prior knowledge from the first module while systematically expanding students' understanding.

Students engage in a classification task, based on constructivist learning principles, deriving models by observing object attributes and applying them to classify new entities, such as a fictional Monster family \cite{Capecchi2}. Initially, students are expected to classify new members based on salient individual features like hairstyle or ear shape \cite{Capecchi2}. Through guided reflection, they recognize the inconsistencies of single-feature classification and refine their strategies by incorporating multiple distinguishing characteristics, leading to a more structured, frequency-based approach. 
After identifying classification features, students explore rule-based models: using a single characteristic as a predictor or a threshold-based approach, where a creature belongs to the Monster family if it exhibits more than half of the defining traits. This leads to a discussion on \textbf{accuracy evaluation}, where students assess their models by comparing predictions with known examples. They explore how different classification rules impact model performance and generalization to unseen data. By testing their models on an unknown dataset, students identify cases where rules fail, introducing the concept of \textbf{overfitting}.

The session ends with a reflection on the real-world applications of AI-based classification systems. Rule-based models provide a structured approach to decision-making but often require data-driven refinement.

\subsection{Module 3: Classification Representations (2 hours)}
The third module introduces structured classification, focusing on how models implement rule-based decision-making to categorize objects. Students expand upon classification concepts by means of multiple representational models.

\begin{figure}[!t]
    \centering
    \begin{subfigure}{0.38\textwidth}
        \centering
        \includegraphics[width=\linewidth]{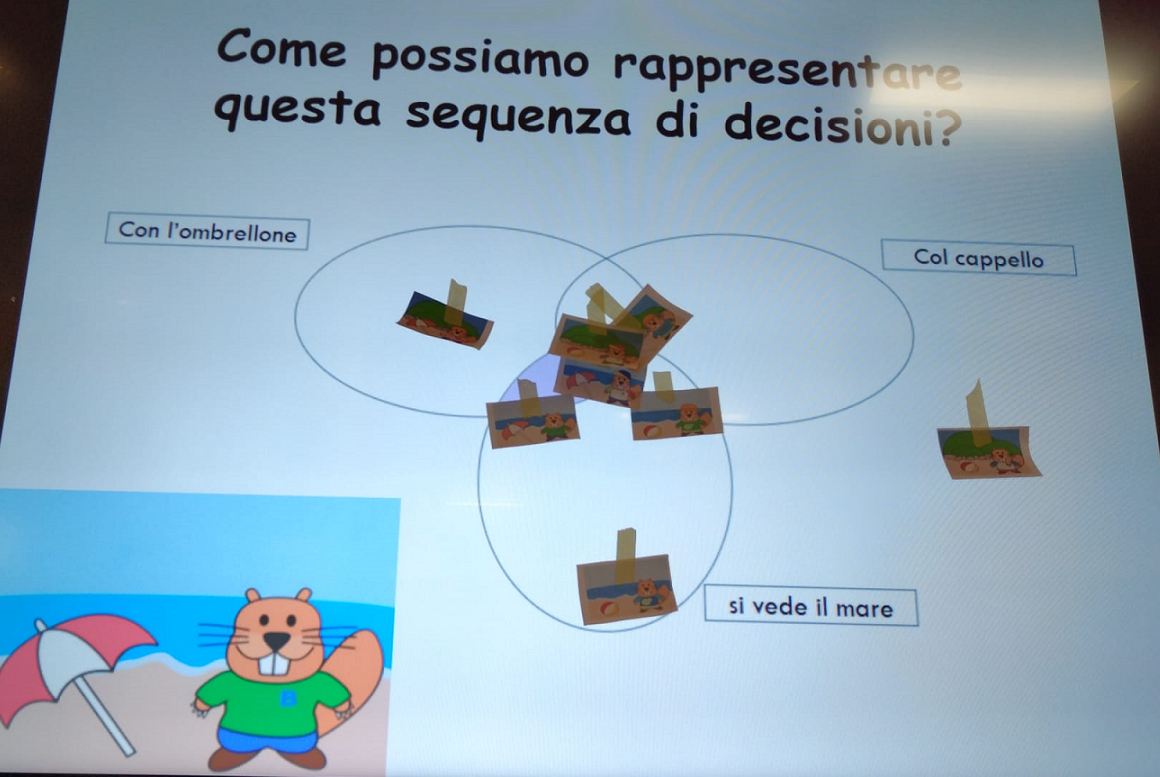}
        \vspace{-4mm}
        \caption{Eulero-Venn diagrams}
        \label{fig:ct-ev}
    \end{subfigure}
    \hfill
    \begin{subfigure}{0.20\textwidth}
        \centering
        \includegraphics[width=\linewidth]{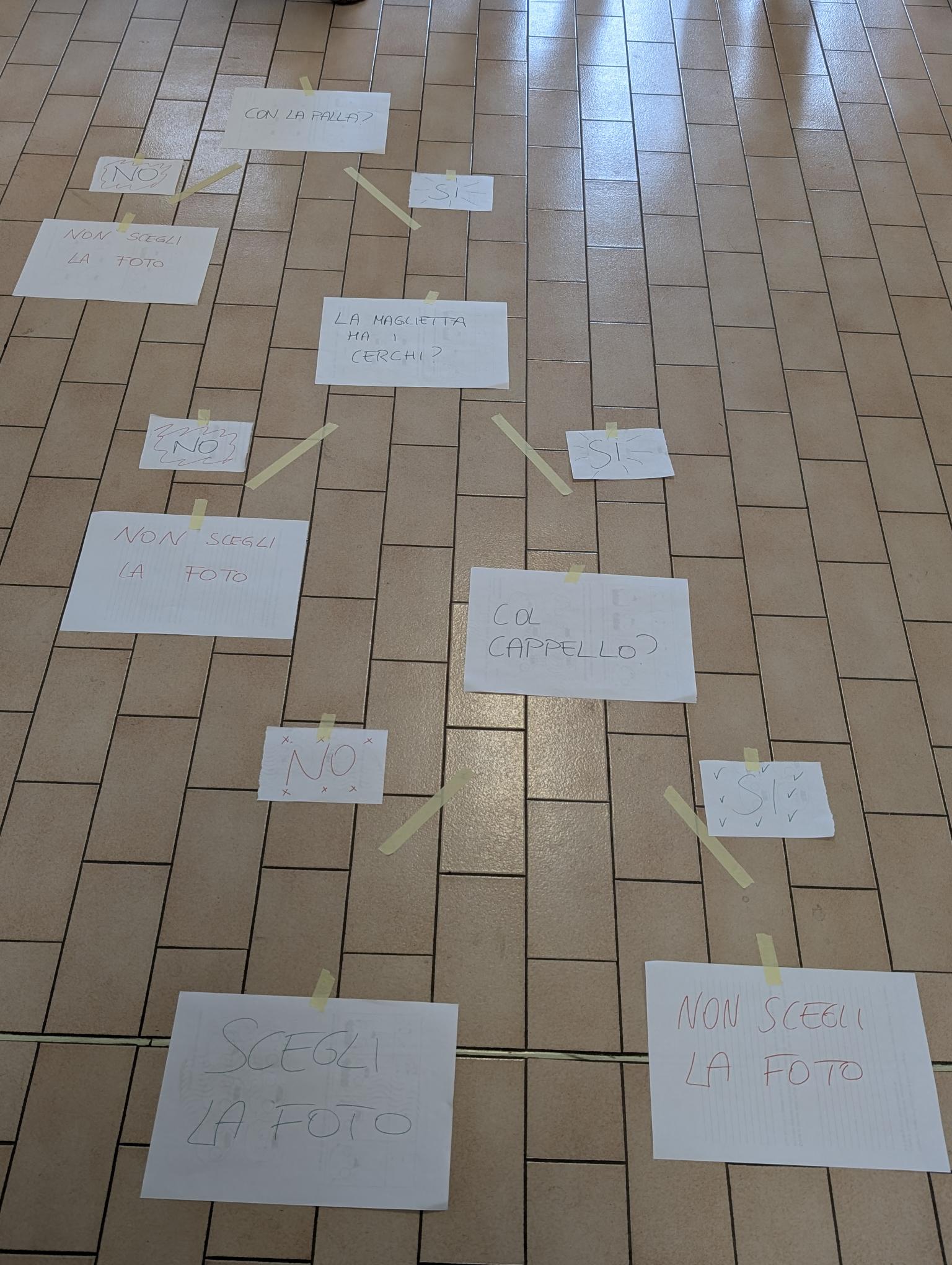}
        \vspace{-4mm}
        \caption{Decision tree}
        \label{fig:ct-dt}
    \end{subfigure}
    \vspace{-2mm}
    \caption{Collaborative classification task [Module 3].}
    \label{fig:colab-task}
\end{figure}

The lesson begins with a discussion on classification principles using two classes, poisonous vs. non-poisonous mushrooms as an example of feature-based differentiation. Students examine how specific characteristics, such as cap color, gill structure, and stem shape, can be used to define classification rules. To reinforce these ideas, they collaboratively construct three distinct classification representations: Euler-Venn diagrams allows to group mushrooms based on shared characteristics; Tabular representations provide a structured data format to highlight classification flexibility; Decision trees illustrate how sequential questions could guide classification decisions. The passage through different semiotic representations develop students understanding \cite{Radford}. Students then apply classification knowledge to new tasks, such as selecting one of eight photos based on predefined rules\footnote{http://kangourou.di.unimi.it/2015/libretto2015.pdf}. This leads to a collaborative activity where students explore problem-solving strategies: brute force analysis, Euler-Venn diagrams (organizing elements into sets on the blackboard (Fig. \ref{fig:colab-task}$_a$)), tabular representations (explicitly recording negative conditions), and decision trees (structuring reasoning hierarchically). This learning-by-necessity approach \cite{Sinha} helps students to experience firsthand the limitations of certain classification models and explore alternative representations. Following CS unplugged principles \cite{Bell,Cs}, students physically engage with decision trees by navigating a floor-based version, following classification paths dictated by rules and monitored by peers (Fig. \ref{fig:colab-task}$_b$).

To deepen their understanding, a more complex fish classification activity is proposed, where students have to determine whether a fish belongs to the poisonous or non-poisonous category based on distinguishing morphological features such as fin shape, eye characteristics, and body color. Students have formulate classification rules and examine different strategies for structuring their decisions. The activity requires to identify common and distinguishing features among fish species, represent classification strategies using tables and Euler-Venn diagrams, and construct a decision tree where each node represents a key distinguishing feature leading to a classification outcome.

As part of the final discussion, proper examples are given to make students reflects on the risks of misclassification, determining the presence of false positives and false negatives and analyzing the implications on real AI-driven systems like facial recognition, autonomous vehicles, and medical diagnostics, where errors are critical. This activity reinforces the importance of robust classification.

\subsection{Module 4: Final Assessment \& Reflection (2 hours)}
The final module is dedicated to evaluating students' learning outcomes and gathering feedback on their overall experience and perceptions.

To evaluate student engagement and their overall perception of the learning path, a satisfaction questionnaire is administered\footnote{The satisfaction questionnaire is administered first to ensure that any difficulties encountered in solving the post-test exercises did not influence students' responses.}. The first set of questions employs a Likert scale to measure students' level of enjoyment, the perceived difficulty of the activities, and their level of comfort while performing the tasks. The subsequent section focuses on individual exercises covered in the previous modules, prompting students to rate tasks 
in terms of difficulty. Students are also asked to indicate the activity they found most engaging and the one they considered least interesting. To capture qualitative insights, the questionnaire included open-ended questions, allowing students to describe two key takeaways from the learning experience and provide additional comments or suggestions. This final reflection enables a comprehensive evaluation of the learning path, considering both cognitive, emotional and perceptive dimensions.

To assess learning effectiveness, a post-test is administered, comprising seven exercises designed to evaluate key computer science concepts (exercises 1-3, 5, 7) and the underlying mathematical skills (exercises 4 and 6). Specifically, the first exercise (\texttt{AI Scenarios}) presented AI-powered objects performing various tasks, prompting students to reflect on whether AI could make mistakes or experience emotions, thereby assessing their understanding of AI’s limitations. The second exercise (\texttt{AI Terms}) involved a fill-in-the-blank task, testing students’ knowledge of AI terminology and key concepts along the AI pipeline. The others are taken from Bebras \cite{Bebras} and from an \textit{anonymized country} standardized test 
and sometimes modified ad hoc, with the aim to verify the ability to use the new knowledge in a different, authentic context \cite{Frezza}.
The third exercise (\texttt{Animal Footprint}) was the same as the ice breaking test of Module 1 in order to verify the improvements in distinguishing characteristics, and explaining classification reasoning. The fourth (\texttt{Frequencies}) focused on decision trees, asking students to analyze a structured decision-making process and complete a corresponding table. The fifth exercise (\texttt{Beaver Structure}) assessed logical sequencing skills, challenging students to follow directional commands to navigate a structured path. The sixth (\texttt{Eulero-Venn}) involved visual data representation, requiring students to determine the truthfulness of statements based on the correct interpretation of an Eulero-Venn diagram. The final exercise (\texttt{Beaver Head}) tested students to identify the correct category based on predefined face features.

\section{Experimental Evaluation}\label{sec:results}
To evaluate our learning path, we investigate three research questions: its impact on students' understanding of AI concepts (\texttt{RQ$_1$}), its intrinsic role in fostering mathematical skills (\texttt{RQ$_2$}), and students' engagement and interest (\texttt{RQ$_3$}). 

\subsection{Experimental Context}

\noindent \textbf{Educational Target.}
Participants were 31 fifth-grade students (35\% female) from two classes in anonymized country, engaged simultaneously in all activities following an open-class model. Among them, 35\% had special educational needs \footnote{Due to privacy constraints, specific details regarding the nature and distribution of learning disorders and disabilities were unavailable, preventing the design of differentiated learning activities and a stratified analysis of results.}, receiving support from specialized teachers as per national regulations. The four modules were delivered over four days during curricular hours by a university math professor with long expertise in K-12 computer science education.

A preliminary meeting with mathematics teachers revealed that students were familiar with fractions, frequency of events, and set operations, though these topics were not covered in current year curriculum. They had basic problem-solving skills but little structured exposure to digital devices, with no formal instruction in programming or computer science. Their only documented CS-related activities included a cryptography and steganography workshop two years earlier and recent participation in Bebras \cite{Bebras}.

All data collection instruments ensured full anonymity, identifying participants by numerical IDs. Of the 25 students (80\% of the original group) present for the final module, 23 (92\%) completed both the satisfaction questionnaire and post-test. Nine out of 11 students with special educational needs participated in the final assessment. Among the 23 respondents, 17 (74\%) attended all modules, four (17\%) participated in three, and two (9\%) attended only two.

\vspace{1mm} \noindent \textbf{Post-Test Scoring Protocol.}
The post-test exercises were evaluated by three independent experts: the mathematics university professor who delivered the activities, a computer science university professor specializing in AI, and a doctoral student in mathematics didactics. Each assessed correctness based on predefined criteria tailored to each exercise (see our repository). For exercises requiring justification (e.g., \texttt{Animal Footprint} and \texttt{Beaver Structure}), evaluators independently scored both correctness and reasoning. Multiple-choice and single-answer questions (e.g., \texttt{AI Terms} and \texttt{Beaver Head}) were verified against an answer key, while structured exercises (e.g., \texttt{Eulero-Venn}) allowed for partial credit based on logical steps. Initially, evaluators scored a subset independently to align grading standards before assessing the remaining responses, with periodic checks.

\subsection{Impact on AI Concepts Understanding [RQ1]}

\begin{figure}[!b]
    \centering
    \begin{subfigure}{0.39\textwidth}
        \centering
        \includegraphics[width=\linewidth]{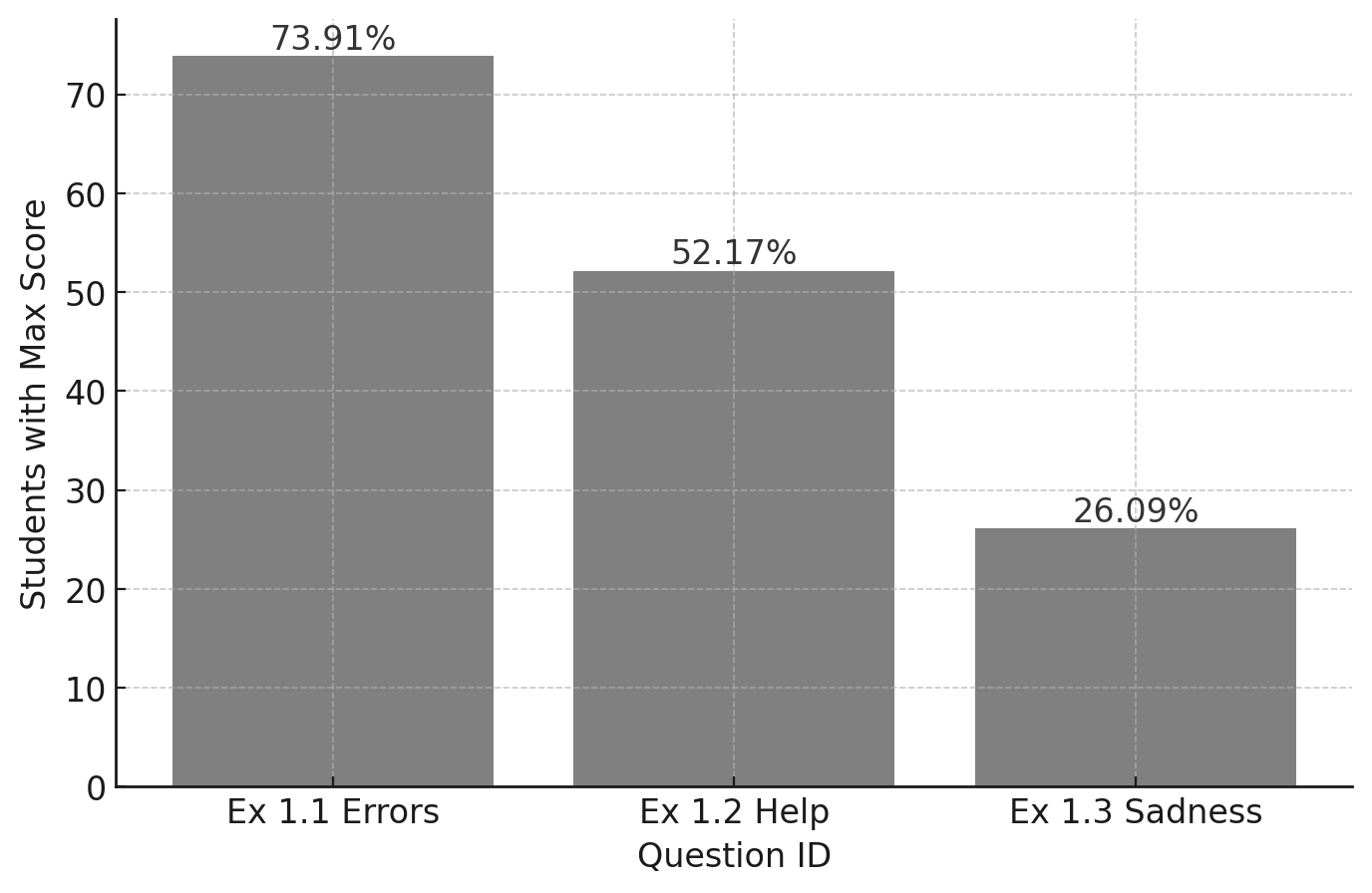}
        \vspace{-6mm}
        \caption{Ex 1: AI Scenarios}
        \label{fig:rq1-ex1}
    \end{subfigure}
    \hfill
    \begin{subfigure}{0.39\textwidth}
        \centering
        \includegraphics[width=\linewidth]{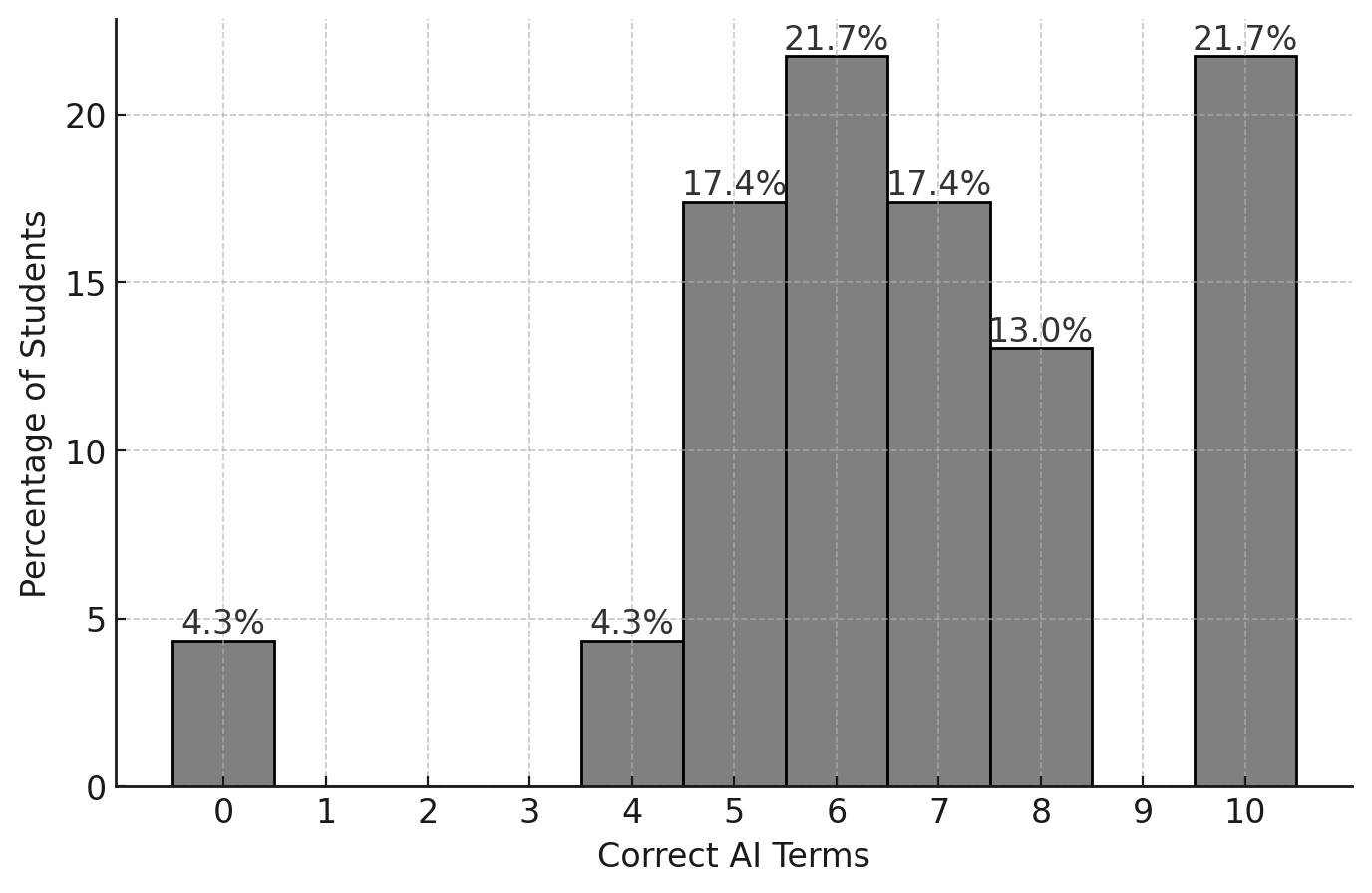}
        \vspace{-6mm}
        \caption{Ex 2: AI Terms}
        \label{fig:rq1-ex2}
    \end{subfigure}
    \hfill
    \begin{subfigure}{0.32\textwidth}
        \centering
        \includegraphics[width=\linewidth]{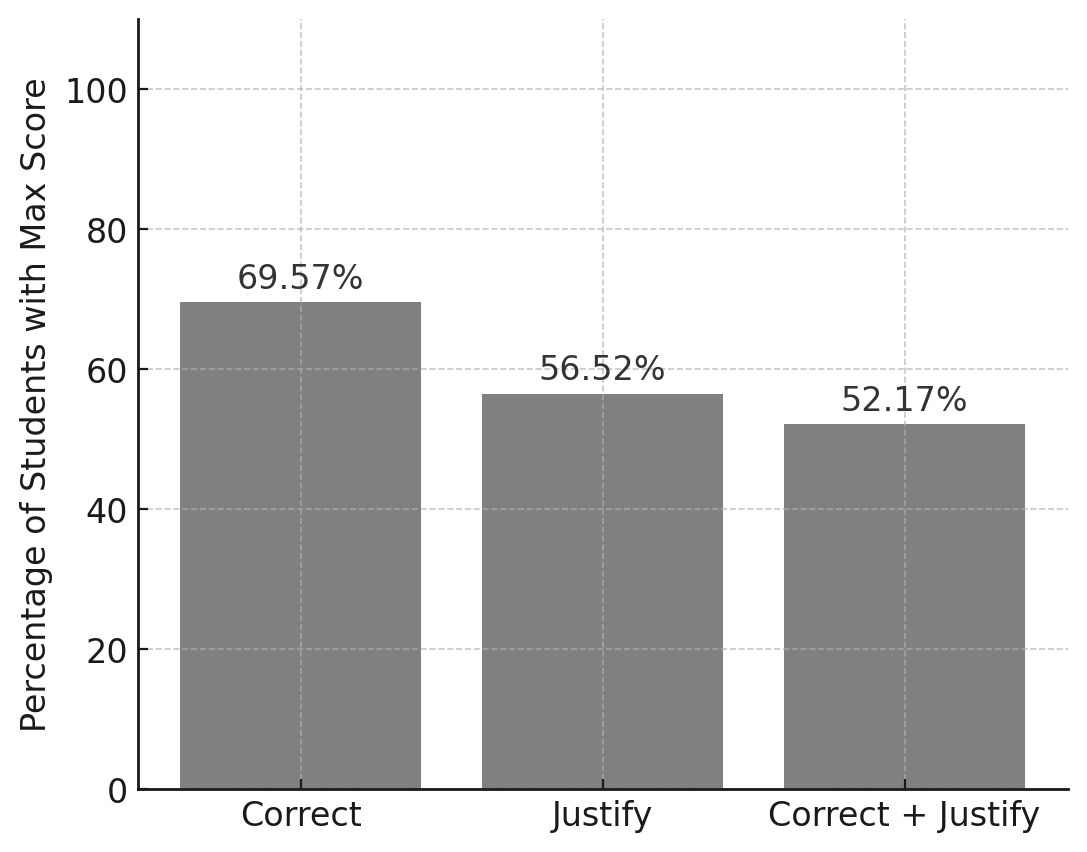}
        \vspace{-4mm}
        \caption{Ex 3: Animal Footprint}
        \label{fig:rq1-ex3}
    \end{subfigure}
    \hfill
    \begin{subfigure}{0.32\textwidth}
        \centering
        \includegraphics[width=\linewidth]{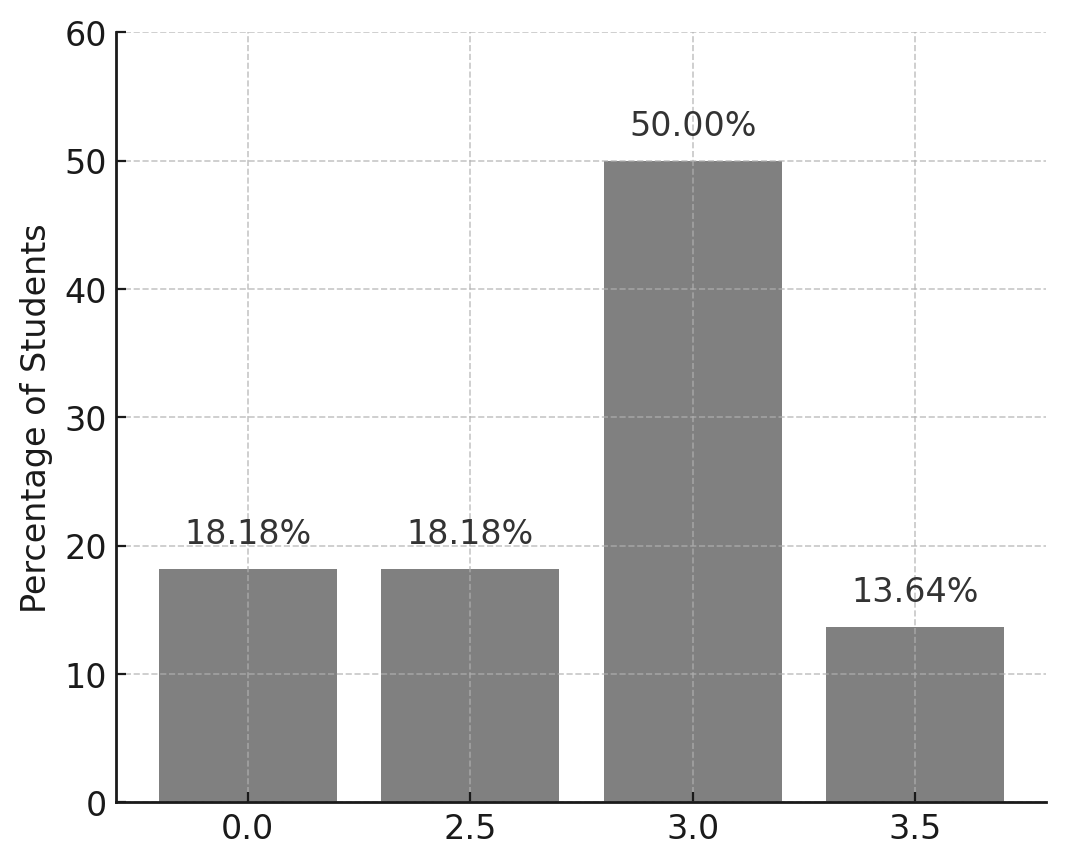}
        \vspace{-4mm}
        \caption{Ex 5: Beaver Structure}
        \label{fig:rq1-ex5}
    \end{subfigure}
    \hfill
    \begin{subfigure}{0.32\textwidth}
        \centering
        \includegraphics[width=\linewidth]{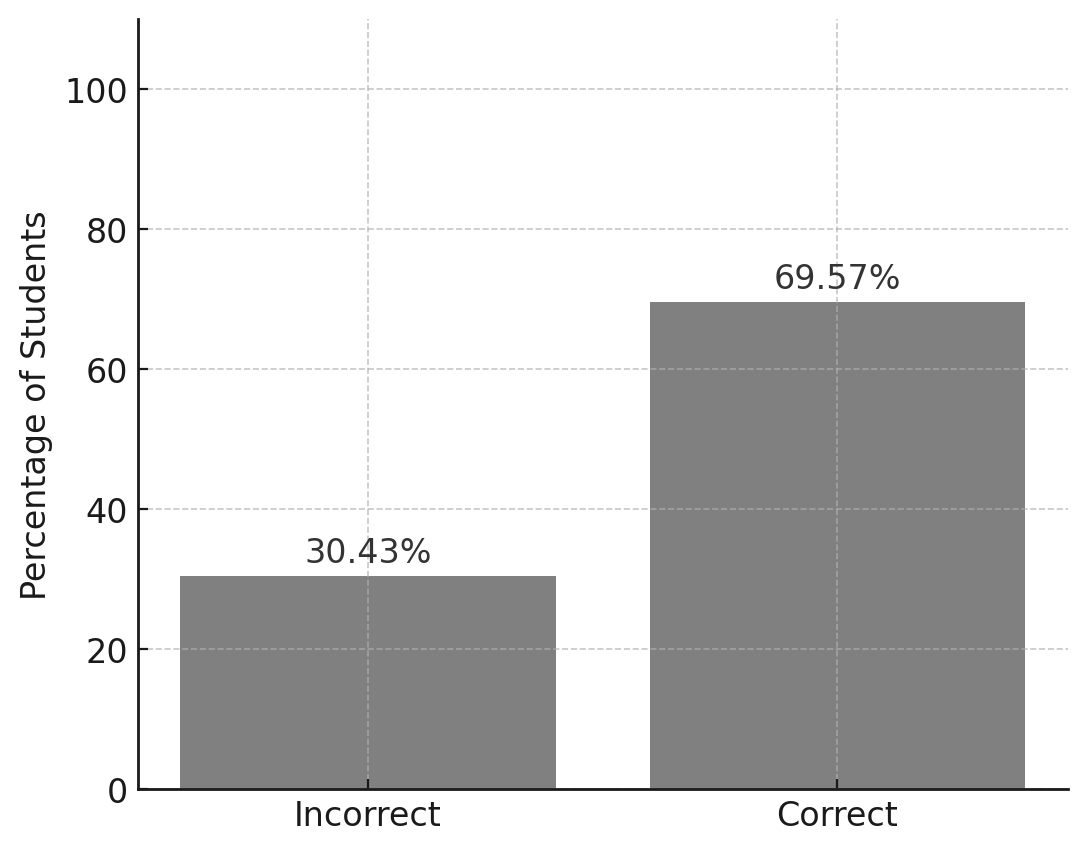}
        \vspace{-4mm}
        \caption{Ex 7: Beaver Head}
        \label{fig:rq1-ex7}
    \end{subfigure}
    \vspace{-2mm}
    \caption{\textbf{[RQ1]} Performance distribution on AI-related post-test exercises.}
    \label{fig:rq1}
\end{figure}

We evaluated students' understanding of AI-related concepts through post-test exercises, measuring their accuracy and reasoning skills. Students performed well in identifying and discussing AI-related errors (Fig.~\ref{fig:rq1-ex1}), with 73.91\% correctly recognizing issues and 26.09\% reflecting on affective states, indicating potential for further development. In understanding AI terms (Fig.~\ref{fig:rq1-ex2}), most students scored between 6 and 10 correct terms, with only 4.3\% scoring 0 or 4, suggesting a solid foundation with room for refinement.

The ability to justify responses (Fig.~\ref{fig:rq1-ex3}) improved significantly, with 69.57\% answering correctly and 56.52\% providing valid justifications, compared to 9.5\% in the initial questionnaire. In structured problem-solving (Fig.~\ref{fig:rq1-ex5}), 50.00\% achieved a strong intermediate level (score of 3.0), while 13.64\% excelled with the highest score (3.5), showing a solid grasp of the concepts. Finally, in classification tasks (Fig.~\ref{fig:rq1-ex7}), 69.57\% provided accurate responses despite the high difficulty level.

\vspace{-1mm}
\hlbox{Answer to RQ1}{Students demonstrated an overall solid understanding of AI concepts, with many achieving high scores across exercises.}

\subsection{Impact on Underlying Mathematical Skills [RQ2]}

\begin{figure}[!b]
    \centering
    \begin{subfigure}{0.4\textwidth}
        \centering
        \includegraphics[width=\linewidth]{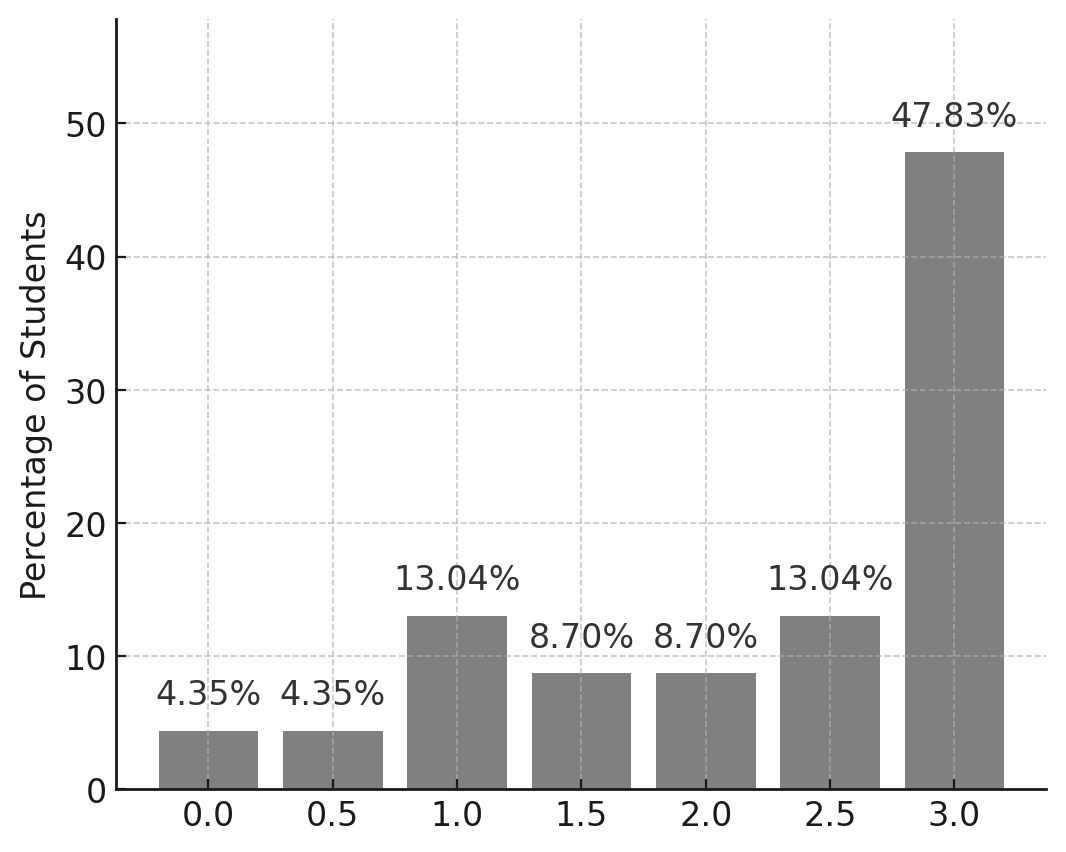}
        \vspace{-4mm}
        \caption{Ex 4: Frequencies}
        \label{fig:rq2-ex4}
    \end{subfigure}
    \hfill
    \begin{subfigure}{0.4\textwidth}
        \centering
        \includegraphics[width=\linewidth]{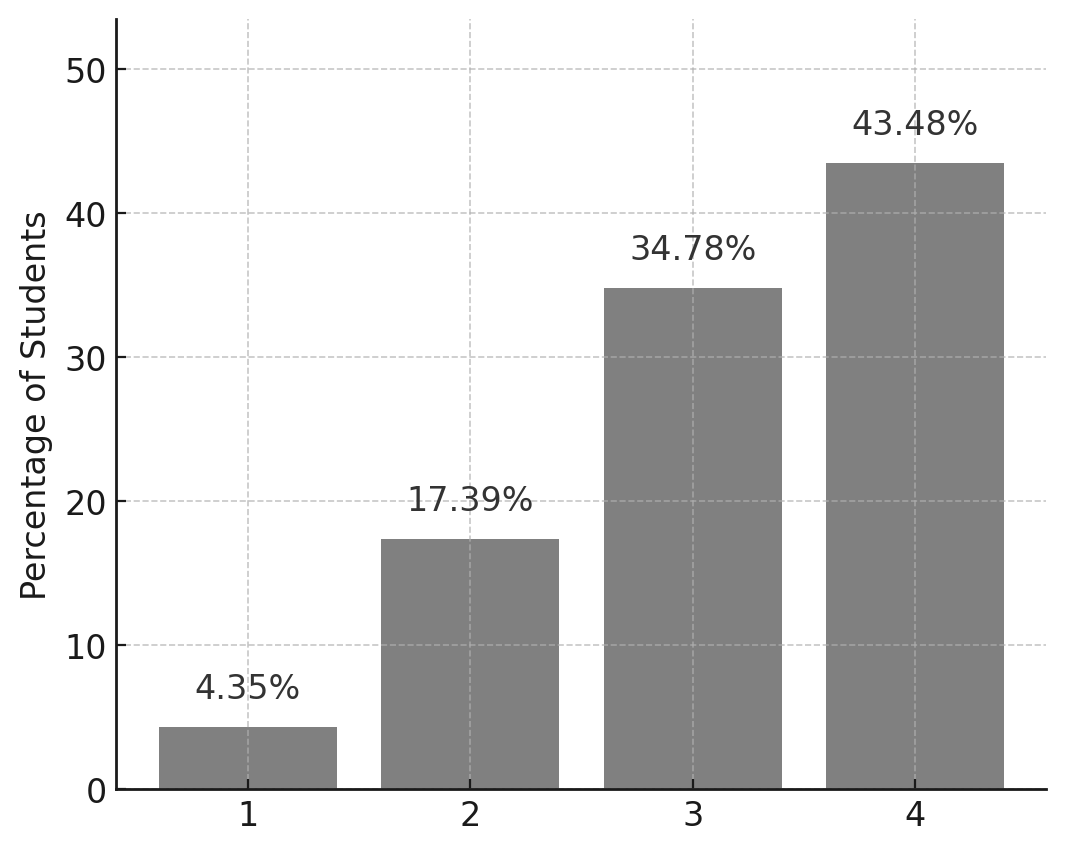}
        \vspace{-4mm}
        \caption{Ex 6: Eulero-Venn}
        \label{fig:rq2-ex6}
    \end{subfigure}
    \vspace{-2mm}
    \caption{\textbf{[RQ2]} Performance distribution on math-related post-test exercises.}
    \label{fig:rq2}
\end{figure}


We evaluated students' mathematical reasoning skills in AI-related tasks, focusing on their ability to interpret structured information, analyze numerical relationships, and apply logical deductions (Fig.~\ref{fig:rq2}). 

Students showed strong skills in interpreting frequency distributions (Fig.~\ref{fig:rq2-ex4}), with 47.83\% achieving the highest score (3) and 13.04\% scoring 2.5, demonstrating proficiency in identifying numerical patterns. Lower scores were more dispersed, with 8.70\% scoring 2, 13.04\% scoring 1, and only 4.35\% below 0.5, suggesting that while most students grasped frequency-based reasoning, some needed further guidance. In set theory tasks (Fig.~\ref{fig:rq2-ex6}), 43.48\% scored full marks and 34.78\% earned a 3, indicating strong logical structuring skills. A smaller group (17.39\%) scored 2, while only 4.35\% received the minimum score (1), suggesting that while most handled set operations well, some may need reinforcement in formal mathematical abstraction.

\vspace{-1mm}
\hlbox{Answer to RQ2}{Students showed a satisfactory grounding in mathematical reasoning, esp. in frequency interpretation and set-based problem-solving.}

\subsection{Impact on Students' Engagement and Interest [RQ3]}

\begin{figure}[!b]
    \centering
    \begin{subfigure}{0.32\textwidth}
        \centering
        \includegraphics[width=\linewidth]{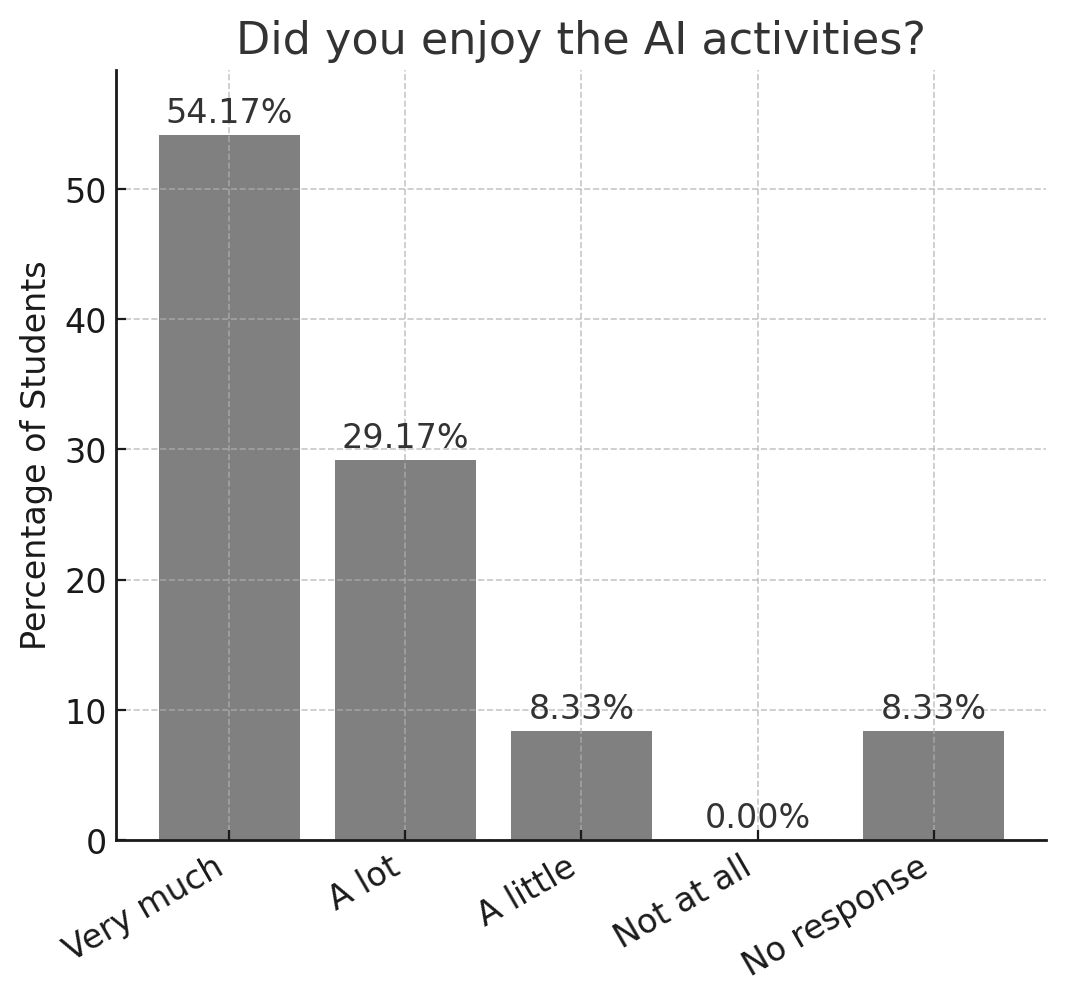}
        \vspace{-4mm}
        \caption{Q1: Enjoyment}
        \label{fig:rq3-q1}
    \end{subfigure}
    \hfill
    \begin{subfigure}{0.32\textwidth}
        \centering
        \includegraphics[width=\linewidth]{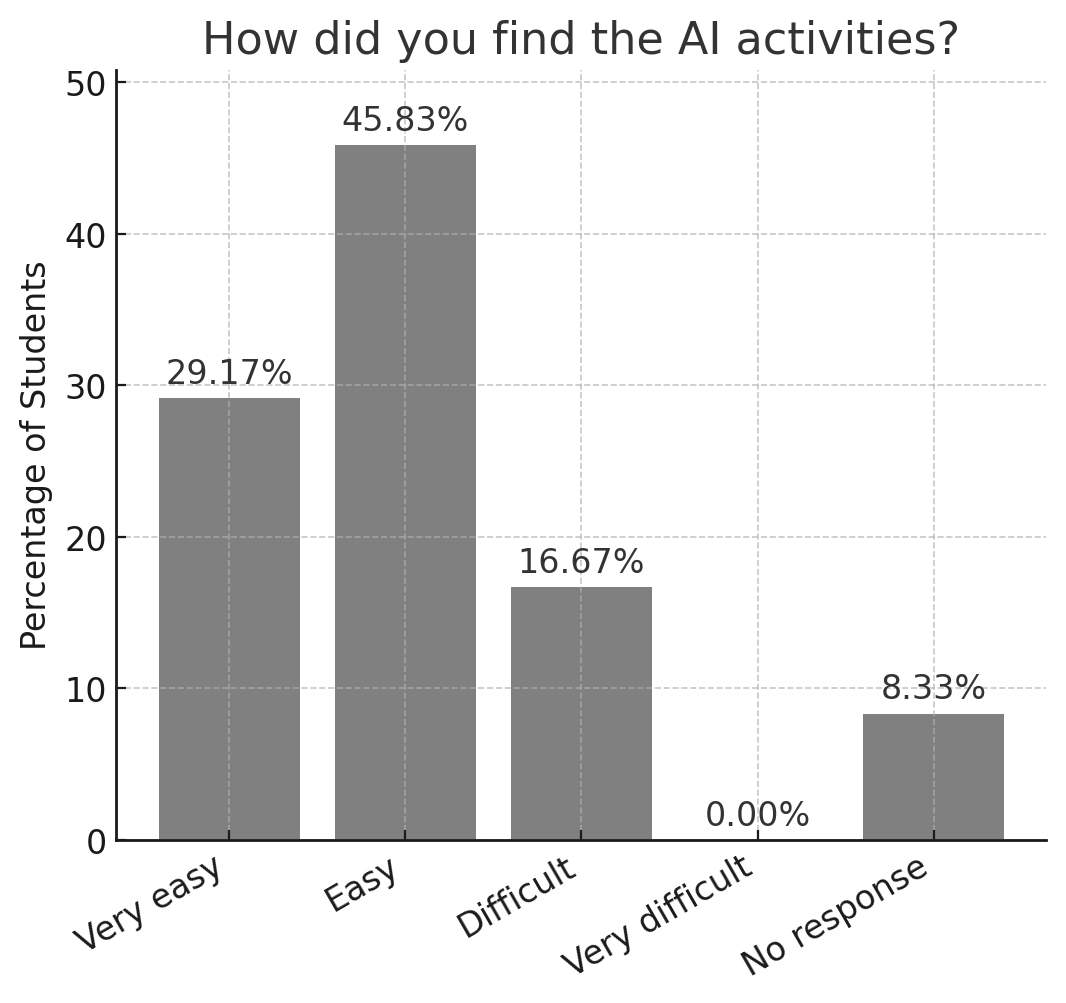}
        \vspace{-4mm} 
        \caption{Q2: Overall difficulty}
        \label{fig:rq3-q2}
    \end{subfigure}
    \hfill
    \begin{subfigure}{0.32\textwidth}
        \centering
        \includegraphics[width=\linewidth]{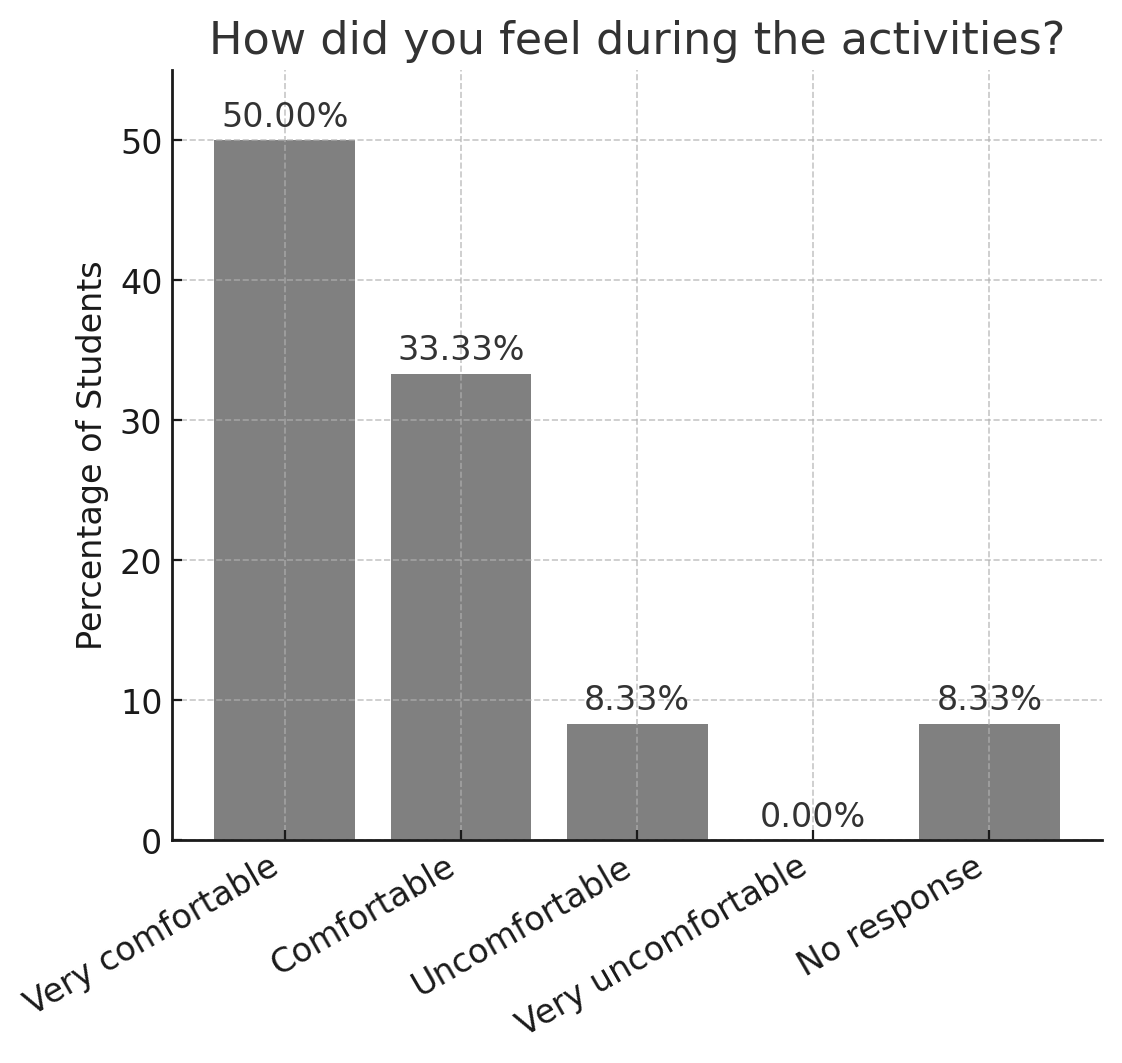}
        \vspace{-6mm}
        \caption{Q3: Feeling}
        \label{fig:rq3-q3}
    \end{subfigure}
    \begin{subfigure}{0.49\textwidth}
        \centering
        \includegraphics[width=\linewidth]{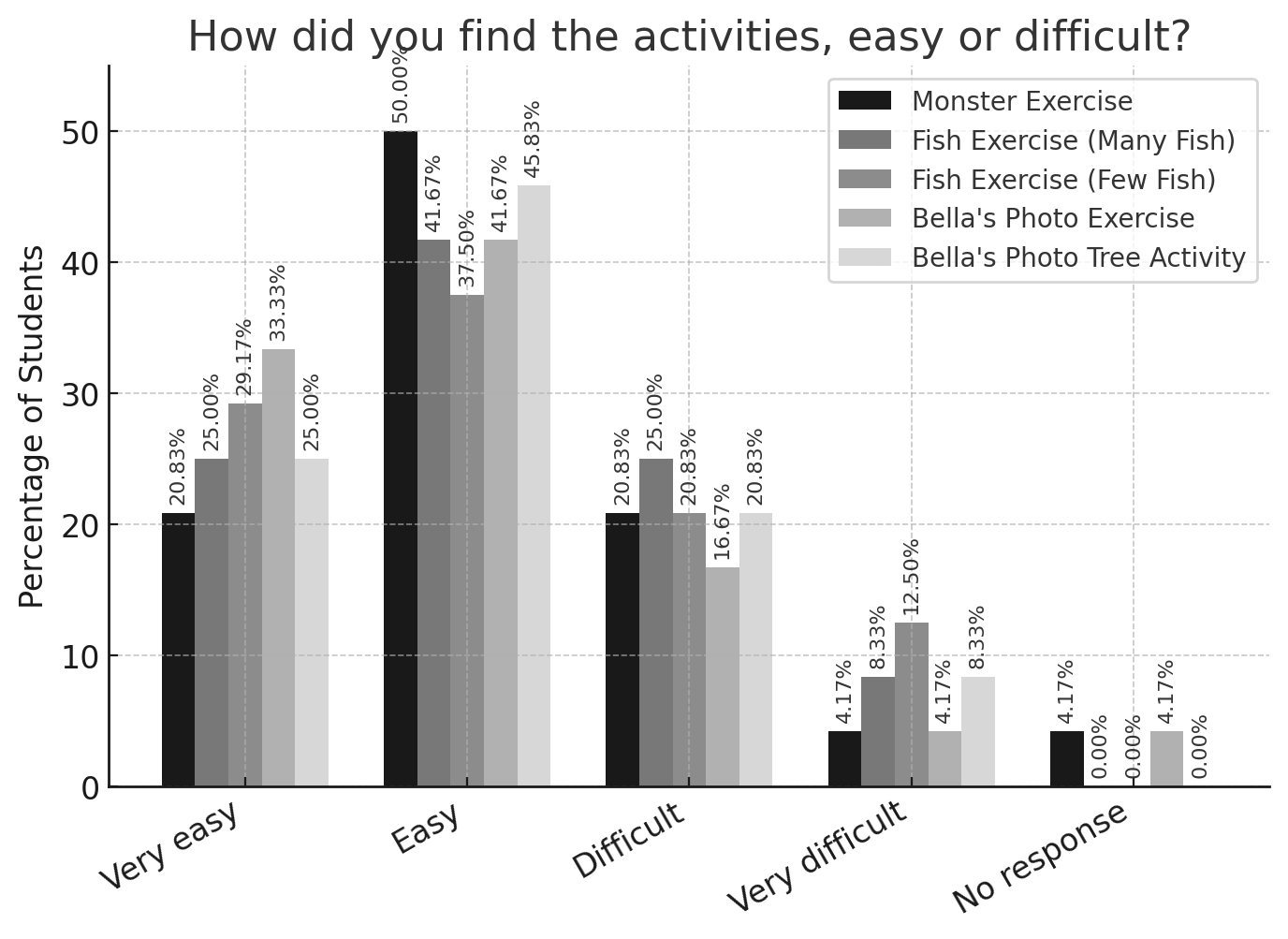}
        \vspace{-6mm}
        \caption{Q4: Activity difficulty}
        \label{fig:rq3-q4}
    \end{subfigure}
    \hfill
    \begin{subfigure}{0.43\textwidth}
        \centering
        \includegraphics[width=\linewidth]{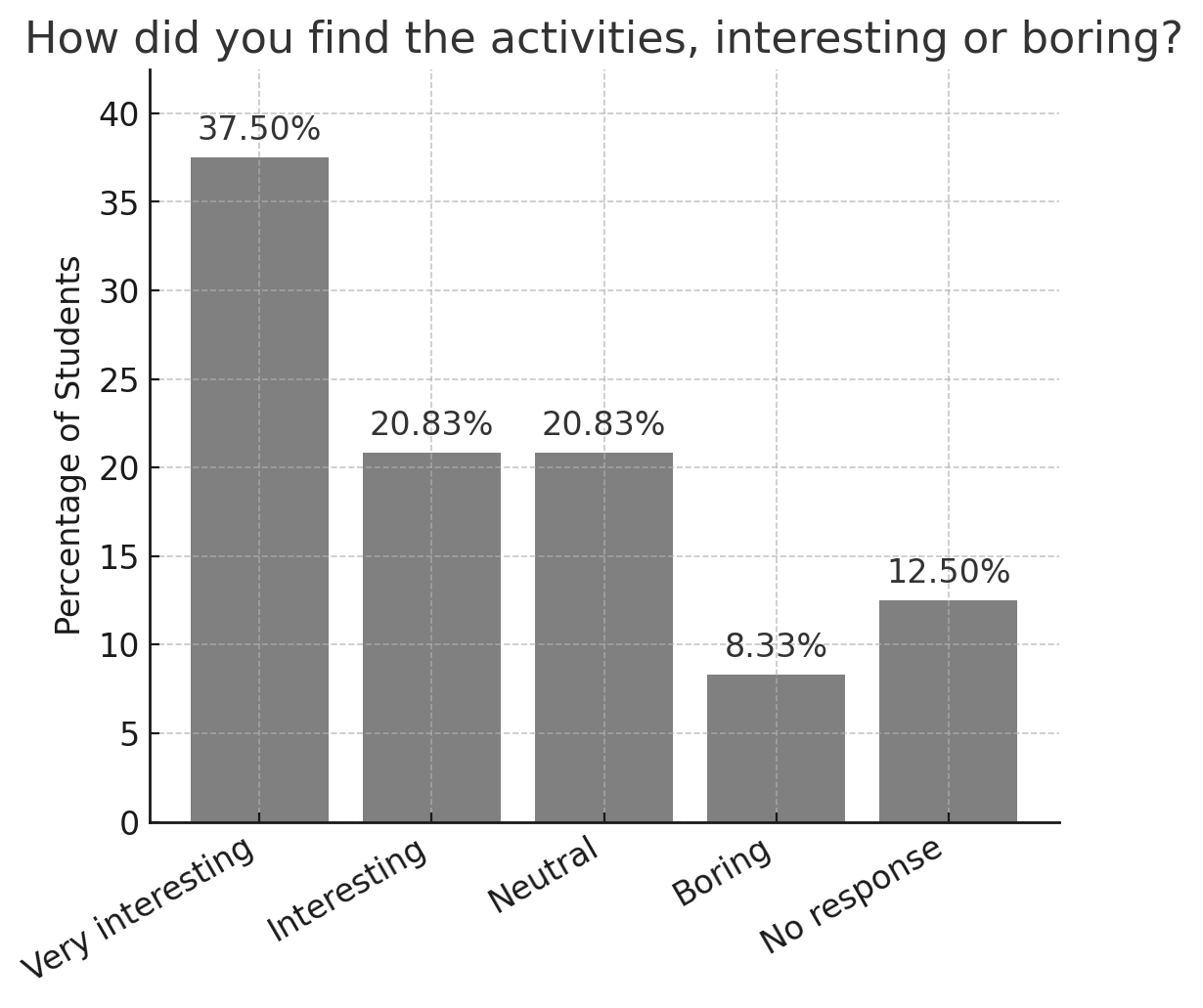}
        \vspace{-6mm}
        \caption{Q5.1: Interest}
        \label{fig:rq3-q5}
    \end{subfigure}
    \hfill
    \begin{subfigure}{0.32\textwidth}
        \centering
        \includegraphics[width=\linewidth]{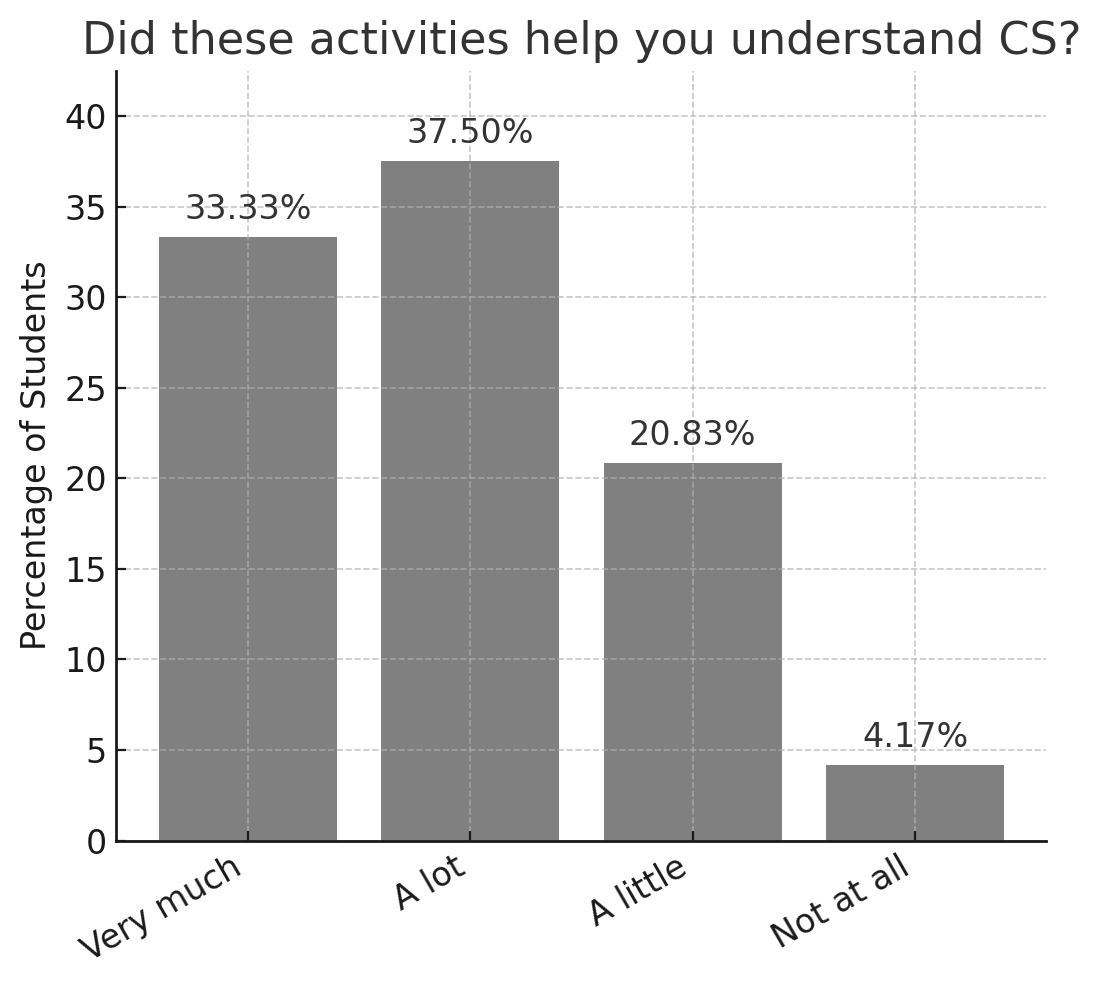}
        \vspace{-6mm}
        \caption{Q6: Understand CS}
        \label{fig:rq3-q6}
    \end{subfigure}
    \hfill
    \begin{subfigure}{0.32\textwidth}
        \centering
        \includegraphics[width=\linewidth]{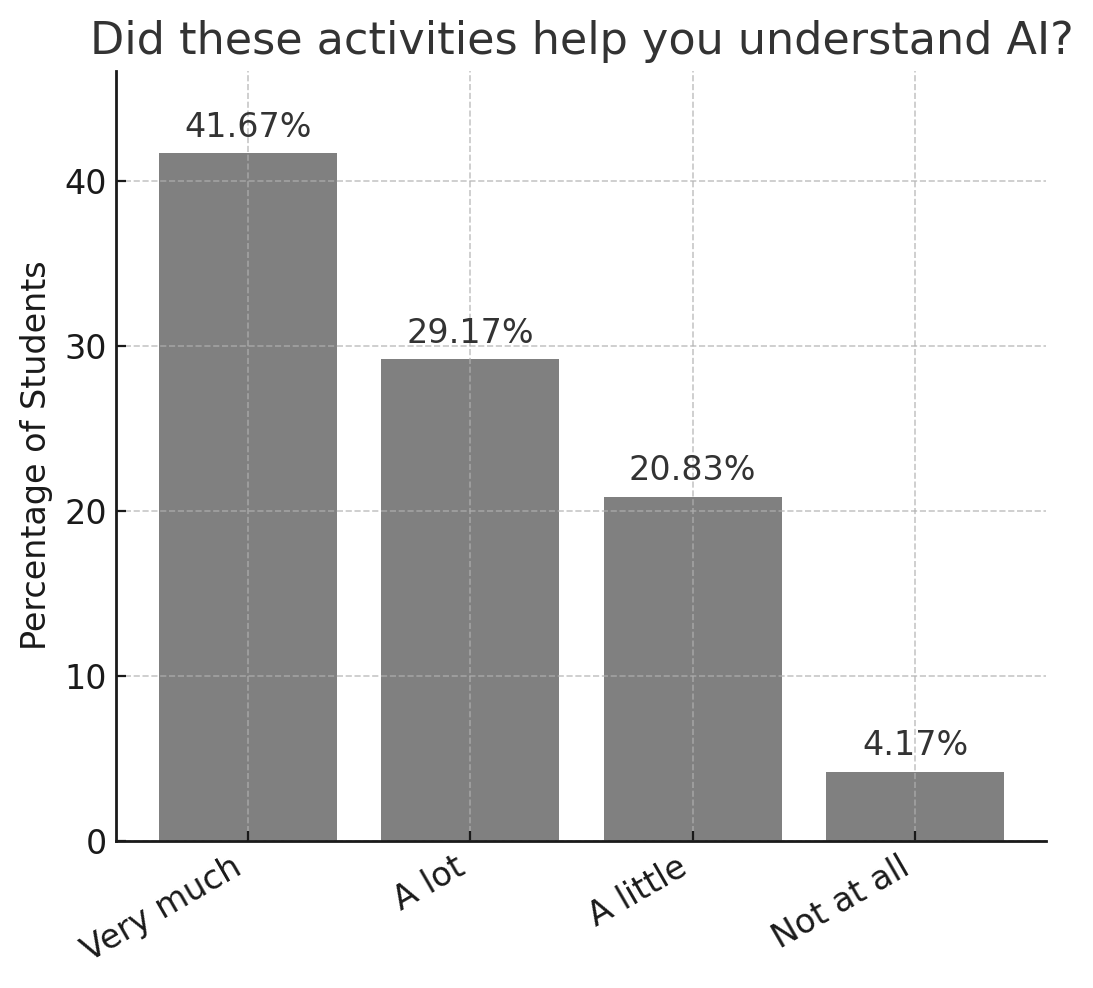}
        \vspace{-6mm}
        \caption{Q7: Understand AI}
        \label{fig:rq3-q7}
    \end{subfigure}
    \hfill
    \begin{subfigure}{0.33\textwidth}
        \centering
        \includegraphics[width=\linewidth]{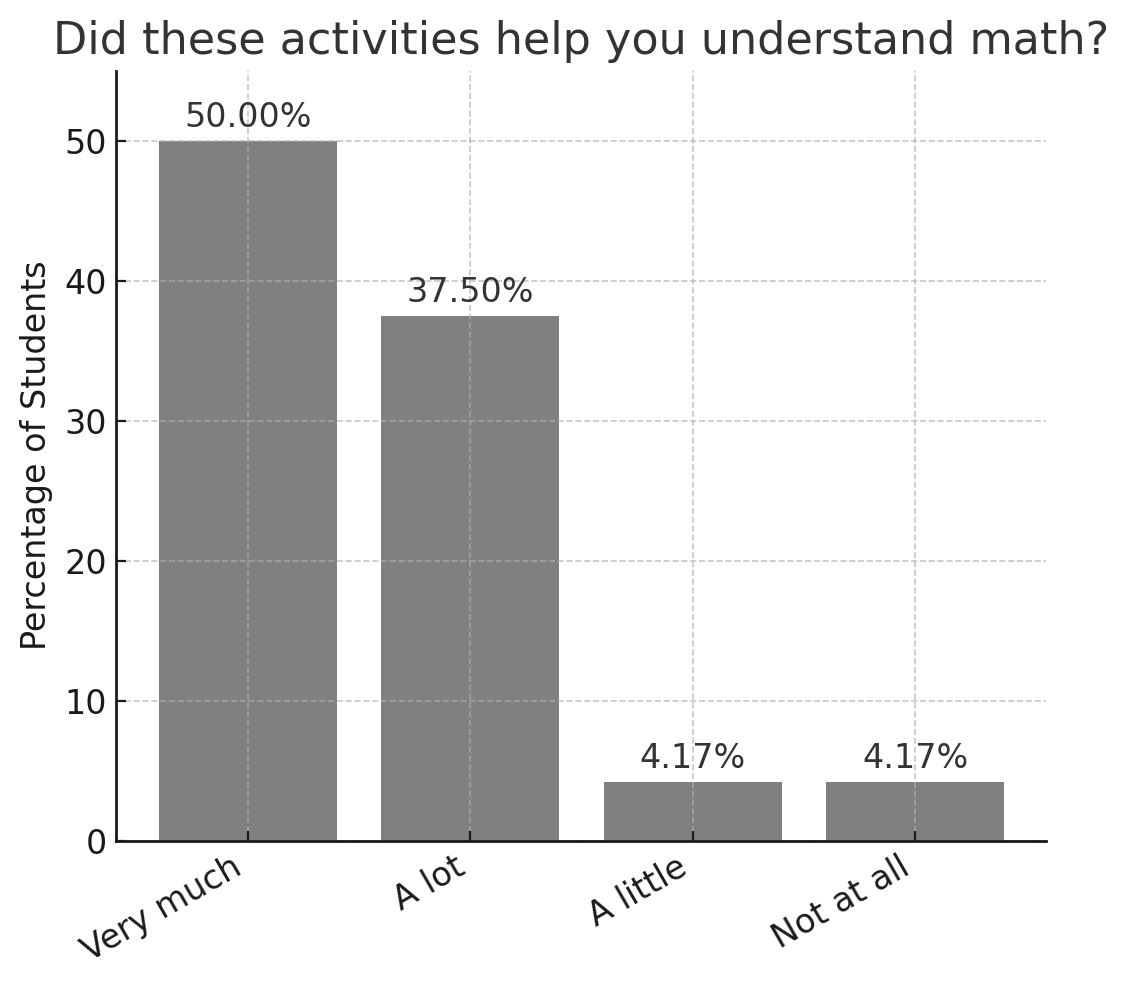}
        \vspace{-6mm}
        \caption{Q8: Understand Math}
        \label{fig:rq3-q8}
    \end{subfigure}
    \vspace{-2mm}
    \caption{\textbf{[RQ3]} Student answers about perceptions of engagement.}
    \label{fig:rq3}
\end{figure}

We assessed how the proposed activities influenced students' engagement and interest in AI and related concepts. Specifically, we looked at enjoyment, perceived difficulty and relevance of the activities ( Figure~\ref{fig:rq3}). Also students reflected on how the activities improved their understanding of CS, AI, and mathematics.

Results indicate that most students found the activities engaging, with 54.17\% enjoying them "very much" and 29.17\% "a lot" (Fig.\ref{fig:rq3-q1}). While 81.67\% rated the activities as "very easy" or "easy", 16.67\% found them difficult (Fig.\ref{fig:rq3-q2}). Most students felt comfortable, with 83.33\% reporting positive emotions (Fig.\ref{fig:rq3-q3}). In terms of learning outcomes, 41.67\% felt the activities greatly helped in understanding CS, 45.83\% in AI, and 50.00\% in mathematics (Figs.\ref{fig:rq3-q6}-\ref{fig:rq3-q7}).

Qualitative feedback supports these results, with students expressing excitement about learning how AI systems recognize objects, particularly enjoying the Monster classification and AI for Oceans activities. Many appreciated the interactive, problem-solving aspects, with one student noting that "classifying monsters was really, really interesting". Some also recognized the connection between AI and mathematics, highlighting how tree structures improve classification. However, a few students mentioned that some activities felt repetitive.

\hlbox{Answer to RQ3}{The activities were overall successful in fostering engagement and interest. While students generally found the learning path enjoyable and educational, refining certain activities to ensure sustained engagement.}

\section{Discussion and Implications}\label{sec:disc}
In this section, we synthesize the findings from the individual experiments, contextualizing them within prior research and drawing educational implications. 
 
Students demonstrated a strong ability to recognize AI-related errors and classify AI concepts but faced challenges when reasoning about affective states and structured problem-solving (\texttt{RQ$_1$}). While they could identify explicit AI behaviors, implicit decision-making processes were more difficult to grasp. Initial responses showed that students primarily associate AI with robots and technological tools rather than computational principles, a common pattern observed in early AI education. To enhance AI literacy, educational approaches should incorporate structured discussions on AI decision-making and ethics, helping students bridge the gap between perception and computational reasoning. Activities that require structured explanations, such as argumentation tasks or guided reflection exercises, could further support the development of reasoning skills.

Mathematical reasoning in AI-related tasks showed strengthened in argumentation and problem-solving (\texttt{RQ$_2$}), particularly when interpreting patterns and relationships. Difficulties emerged in abstraction (e.g. in substituting the correct values in the definition of accuracy) and many students relied on procedures rather than deductive reasoning. We believe that it is necessary to strengthen that part of the proposal that concerns the choice of classification criteria, supplementing it with completions of various kind of graphs \cite{Grillenberger}.

Finally, students found the activities engaging (\texttt{RQ$_3$}), with problem-solving and interactive tasks being particularly well received. However, some exercises were perceived as repetitive, highlighting the need to balance structured guidance with exploratory learning. Initial responses revealed that misconceptions about AI were common, with students attributing emotions and moral superiority to AI systems, reflecting the influence of media narratives. Addressing these misconceptions explicitly through guided discussions and real-world AI applications could refine understanding. Additionally, students showed strong engagement when AI activities were linked to mathematics, reinforcing the effectiveness of interdisciplinary approaches in sustaining motivation \cite{Sanusi}. 

Overall, the path promoted an improvement in the children's classification and representation skills and found a good level of enjoyment; this last aspect is relevant as one of the critical issues that has emerged in the literature concerns the absence of stimulating activities within the school context \cite{Grillenberger}. The contextualization to the understanding of the functioning of AI increases the interest in mathematical concepts, which in turn facilitates the comprehension of the functioning of the mechanisms underlying AI, in a virtuous circle that strengthens the acquisition of fundamental skills in both disciplines, and above all in the area of conscious citizenship.

\section{Conclusions and Future Work}\label{sec:conc}
In this paper, we proposed a learning path for foundational AI literacy, and we examined how young students engage with AI concepts, mathematical reasoning in AI-related tasks, and their overall interest in is context. Through a structured evaluation combining post-test exercises and qualitative reflections, we found that students demonstrated a strong ability to identify AI errors and classify AI terms, yet encountered challenges in reasoning about implicit AI behaviors and structured problem-solving. Their mathematical reasoning was solid in structured problem-solving and logical analysis but revealed difficulties in abstract generalization and transferring knowledge across different types of representations. Engagement levels were high, particularly in activities that integrated AI with mathematics, reinforcing the call for interdisciplinariety.

Despite these contributions, this study has some limitations. The participant group was relatively small and only involved classes from one school, limiting generalizability, and the study focused on short-term learning outcomes rather than long-term retention. Additionally, while the evaluation framework captured both quantitative and qualitative insights, further research is needed to assess how students' understanding evolves over extended periods. Future iterations should also introduce students to block-based programming environments beforehand, allowing them to directly experience data collection and transformation processes, reinforcing the necessity of labeled data in AI systems. Moreover, considering that progress in thought and language does not always align in mathematical comprehension \cite{Vygotskji}, future implementations should explore how different representational formats influence students' understanding and how to scaffold transitions between them. Finally, expanding this work to middle school students would allow for an adaptation of activities that align with their mathematical background, ensuring that they scale appropriately across different levels.

\end{document}